\documentclass[journal=jacsat,manuscript=article]{achemso}
\usepackage[version=3]{mhchem}
\usepackage[utf8]{inputenc}
\usepackage[T1]{fontenc}
\usepackage{bm}
\usepackage{float}
\usepackage{adjustbox}
\usepackage{xcolor}
\usepackage{booktabs}
\usepackage{multirow}
\usepackage{amssymb}
\usepackage{graphicx}
\usepackage{subfig}
\usepackage{xr}

\makeatletter
\newcommand*{\addFileDependency}[1]{
  \typeout{(#1)}
  \@addtofilelist{#1}
  \IfFileExists{#1}{}{\typeout{No file #1.}}
}
\makeatother

\newcommand*{\myexternaldocument}[1]{
    \externaldocument{#1}
    \addFileDependency{#1.tex}
    \addFileDependency{#1.aux}
}

\myexternaldocument{SI}
\usepackage[parfill]{parskip}
\author{Carter Knutson $^{||}$}
\author{Mridula Bontha $^{||}$}
\author{Jenna A. Bilbrey $^{||}$}
\author{Neeraj Kumar$^{||}$}
\affiliation[BSF]
{$^{||}$
Pacific Northwest National Laboratory, 902 Battelle Blvd,
Richland, WA 99352, United States}

\email{neeraj.kumar@pnnl.gov}
\phone{+1 509-372-6422}

\title{Decoding the Protein-ligand Interactions Using Parallel Graph Neural Networks}

\abbreviations{PLI, GNN}

\begin{document}

\begin{abstract}


Protein-ligand interactions (PLIs) are fundamental to biochemical research and their identification is crucial for estimating biophysical and biochemical properties for rational therapeutic design. Currently, experimental characterization of these properties is the most accurate method, however, this is very time-consuming and labor-intensive. A number of computational methods have been developed in this context but most of the existing PLI prediction heavily depend on 2D protein sequence data. Here, we present a novel parallel graph neural network (GNN) to integrate knowledge representation and reasoning for PLI prediction to perform deep learning guided by expert knowledge and informed by 3D structural data. We develop two distinct GNN architectures: GNN\textsubscript{F} is the base implementation that employs distinct featurization to enhance domain-awareness, while  GNN\textsubscript{P} is a novel implementation that can predict with no prior knowledge of the intermolecular interactions. Comprehensive evaluation demonstrated that GNN can successfully capture the binary interactions between ligand and protein's 3D structure with 0.979 test accuracy for GNN\textsubscript{F} and 0.958 for GNN\textsubscript{P} for predicting activity of a protein-ligand complex. These models are further adapted for regression tasks to predict experimental binding affinities and pIC\textsubscript{50} crucial for drug's potency and efficacy. We achieve a Pearson correlation coefficient of 0.66 and 0.65 on experimental affinity and 0.50 and 0.51 on pIC\textsubscript{50} with GNN\textsubscript{F} and GNN\textsubscript{P}, respectively, outperforming similar 2D sequence based models. Our method can serve as an interpretable and explainable artificial intelligence (AI) tool for predicted activity, potency, and  biophysical properties of lead candidates. To this end, we show the utility of  GNN\textsubscript{P} on SARS-Cov-2 protein targets by screening a large compound library and comparing our prediction with the experimentally measured data. 

\end{abstract}

\flushbottom
\maketitle

\thispagestyle{empty}

\section*{Introduction}
\label{introduction}

Accurate prediction of protein-ligand interactions (PLI) is a critical step in therapeutic design and discovery. These interactions influence various molecular-level properties, such as  substrate  binding, product release, regio-selectivity, target protein function, and ability to facilitate potential hit identification, which is the first step in finding novel candidates for drug discovery \cite{chen2021predicting}. With increases in computing power, code scalability, and advancement of theoretical methods, physics-based computational tools such as molecular dynamics and molecular/quantum mechanics can be used for the reliable representation of PLI and prediction of accurate binding free energies \cite{clyde2021high, wang2017accurate, beierlein2011simple}. However, these methods are computationally expensive and are limited to a number of protein-ligand complexes \cite{sliwoski2014computational}. This limits their routine use in high-throughput virtual screening \cite{wang2017accurate,beierlein2011simple} for the discovery of novel hit candidates and lead optimization \cite{boniolo2021artificial} for a given protein target. Molecular docking has been used to predict binding affinity and estimate interactions with reasonable computational cost \cite{venkatachalam2003ligandfit, allen2015dock, ruiz2014rdock, zhao2013discovery, jain2003surflex, jones1997development, friesner472004, vina2010improving, clyde2021high}; however, its accuracy is relatively low as it uses heuristic rules to evaluate the scoring function.

There has been significant effort to develop deep learning models that predict PLI \cite{zhong2018artificial, wen2017deep}, and other biophysical properties that are critical for therapeutic design but cannot be predicted through physics based modeling \cite{karimi2019deepaffinity}. The greater understanding of PLI enabled by deep learning can help in the estimation of properties such as activity, potency and binding affinity \cite{ozturk2018deepdta}. However, several technical challenges limit the use of deep  learning for modeling protein-ligand complexes and accurate prediction of properties. The first challenge relates to the limited availability of protein-ligand 3D data and the second challenge focuses on the appropriate representation of the data (domain knowledge), specifically in terms of the comprehensive 3D geometry representation. Structure-based methods have the advantage of producing results that can be interpreted but are limited by the number of available samples \cite{karimi2019deepaffinity}. Circular fingerprints, generated by encoding localized structural and geometric information, have long been a cornerstone of cheminformatics \cite{glen2006circular}. The flexibility of fingerprints has created new avenues for molecular computational research, including the increased implementation of graph-based representations to include domain awareness \cite{duvenaud2015convolutional, kearnes2016molecular}. Molecular graph representations provide a way to model and simulate the 3D chemical space while retaining a wider range of structural information. In that context, Ragoza et al.~implemented a deep convolutional neural network (CNN) that operates directly on 3D molecular graph input, similar to the AtomNet model previously implemented by Walloch et al.~\cite{ragoza2017protein, wallach2015atomnet}. Other approaches such as Graph-CNN developed by Torng et al. use unsupervised autoencoders to leverage sequence-based data that is more abundant but also costly in terms of structural accuracy \cite{torng2019graph}.

Graph-based representations extend the learning of chemical data to graph neural networks (GNNs). Monti et al.~designed a mixture model network (MoNet) that enables non-Euclidean data, such as graphs, to be learned by CNNs \cite{monti2017geometric}. That approach has been generalized and improved through the formulation of graph attention (GAT) networks \cite{velivckovic2017graph}.
GAT architectures operate on the importance of a given node, leading to improved computational efficiency and accuracy \cite{li2015gated}. Lim et al.~showed the implementation of such an architecture to capture PLI, which provides a baseline for the development of more robust models \cite{lim2019predicting}. Chen et al.~proposed a bidirectional attention-driven, end-to-end GNN to predict PLI and enable biochemical insights through attention weight visualization \cite{chen2021predicting}. 
Predicting the activity of a protein-ligand complex is a binary classification problem. Reshaping the problem to focus on affinity creates a regression problem of heightened complexity.
The existing deep learning models that predict the binding affinity or other biophysical and biochemical properties such as   IC\textsubscript{50}, K\textsubscript{i}, K\textsubscript{d}, and EC\textsubscript{50} \cite{ozturk2018deepdta} \cite{karimi2019deepaffinity} \cite{li2020monn}. However, most of the methods use sequence-based data for proteins and SMILES representations for interacting ligands. For example, DeepDTA \cite{ozturk2018deepdta} and DeepAffinity \cite{karimi2019deepaffinity} use SMILES strings of the ligands and amino-acid sequences of the target proteins to predict the affinity. MONN \cite{li2020monn} is a multi-objective sequence-based neural network model that first predicts the non-covalent interaction between the ligand and the residues of the interacting target and then the binding affinities in terms of IC\textsubscript{50}, K\textsubscript{i}, and K\textsubscript{d}. Such methods are accessible due to the abundant availability of sequence-based data, but do not capture 3D structural information in the PLI and predicting regression properties. Binding is best understood when the 3D pocket of the target is known, and \textit{in situ}, the protein-ligand complex is formed due to changes in the conformation of the 3D structure of the protein and ligand post-translation.

In this contribution, we formulated two GNNs based on the GAT architecture by incorporating domain-specific featurization of the protein and ligand atoms (GNN\textsubscript{F}) and by implementing parallel GAT layers such that GNN\textsubscript{P} uniquely learns the interaction with limited prior knowledge. The inclusion of different features on the protein and ligand atoms enables our models to be more physics informed. The implementation of GAT layers combined with our featurization enables the model to learn the representation and the chemical space of the training data. We further use these models to predict experimental binding affinity and pIC\textsubscript{50} of the protein-ligand complex. This allows us to leverage the 3D structures of the target protein, ligands, and the interaction between them which is crucial both for the activity and affinity prediction.

\section{Methods}

\subsection{Network Architecture}

The goal in this work is to define a GNN architecture that predicts characteristics of a protein-ligand pair by learning features of the protein and ligand that may not be obvious to the human observer. Our molecular graph structure is defined as G\{V,E,A\}, where V is the atomic node set, E is the corresponding edge set, and A is the adjacency matrix. Given the diverse structural properties of protein-ligand complexes, we include additional biomolecular domain-aware features to previous GAT architectures \cite{lim2019predicting, velivckovic2017graph} by defining distinct featurizations for the protein and ligand components, as shown in Table \ref{tab:features}, denoted GNN\textsubscript{F}, and by removing the dependency on prior knowledge of the protein-ligand interaction through the implementation of parallel GAT layers, denoted GNN\textsubscript{P}.

\begin{figure}[H]
\begin{center}
\centerline{\includegraphics[trim={0 0 0 0cm},clip,width=22cm,height=12cm]{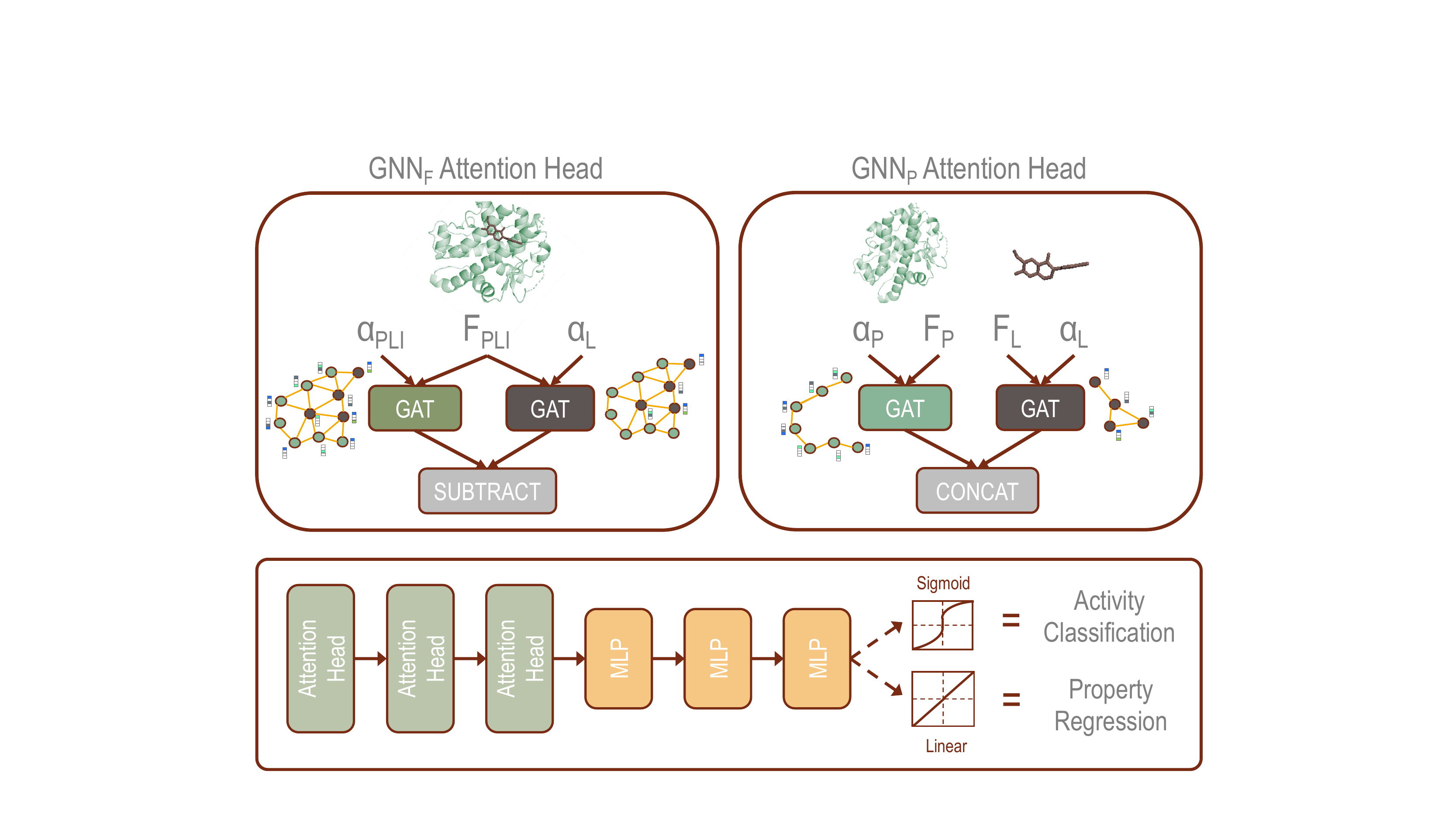}}
\caption{Schematic showing the prediction logic implemented in our GNN models.
The two models differ based on the applied attention head.
GNN\textsubscript{F} uses the PLI obtained from docking simulations to create a combined feature and adjacency matrix. In GNN\textsubscript{P}, the features for the ligand and target protein are coded separately alongside their corresponding adjacency matrices. The output from the attention head is passed through a series of MLP that can be tuned for activity classification through application of the sigmoid activation function and binary cross-entropy loss function or property regression through application of the linear activation function and mean squared error loss function.}
\label{fig:GNNSP}
\end{center}
\end{figure}

The GNN\textsubscript{F} and GNN\textsubscript{P} models differ in the architecture of the attention head as seen in Figure \ref{fig:GNNSP}, a schematic of the prediction logic implemented in our models. In GNN\textsubscript{F}, the protein and ligand adjacency matrices are combined into a single matrix, and edges are added between protein and ligand nodes based on the distance matrix obtained from docking simulations. The GNN\textsubscript{F} attention head uses a joined feature matrix for the ligand and target protein, which is passed into one GAT layer that learns attention based on the PLI adjacency matrix and a second GAT layer that learns attention based on the ligand adjacency matrix. The output of these two GAT layers are subtracted in the final step of each attention head.

In the absence of a co-crystal structure of a protein-ligand complex, docking simulations are typically performed to model the PLI. In GNN\textsubscript{P}, the 3D structures of the protein and ligand are initially embedded separately based on their adjacency matrices, which represent internal bonding interactions. The GNN\textsubscript{P} attention head passes separate features for the protein and ligand to individual GAT layers that learn attention based on the respective adjacency matrix. The outputs of the GAT layers are concatenated in the final step of the attention head. Separation of the ligand and protein in parallel GAT layers preformed by GNN\textsubscript{P} removes prior information about the interactions, providing a foundation to remove the need for docked structural information. This discrete representation enables us to enter the protein and ligand directly into the GNN without knowing the prior protein-ligand interaction, which  otherwise needed to be computed using physics-based simulations. 

In both models, each node is given a set of features \emph{F}, described in Table \ref{tab:features}, which are engineered with an emphasis toward biochemical information, using the molecular Python package RDKit. When input into a GAT layer along with the corresponding adjacency matrix, each node feature is transformed by a learned weight matrix \textbf{\emph{W}} $\mathbb{\in R}^{F \times F}$, where \textit{F} is the dimensions of the node features attributed to input represented by h = \{\^{h\textsubscript{1}}, \^{h\textsubscript{2}},...,\^{h\textsubscript{\textit{N}}}\} where \^{h\textsubscript{\textit{i}}} $\mathbb{\in R}^{F}$ and \textit{N} is the number of atoms.
The attention coefficient $e_{ij}$ for interacting atoms $i$ and $j$ is calculated as a summation of the importance of the $i$-th node interaction with the $j$-th node and vice versa. For node $i$ given an input feature matrix $\hat{h_i}$ at convolution layer $l$, the attention coefficient  $e_{ij}$ is given as: 
\begin{equation}
\label{attention_coefficient_eq1}
    e_{ij} = \hat{h^{\prime}_{i}} \boldsymbol{W} \hat{h^{\prime}_{j}} + \hat{h^{\prime}_{j}} \boldsymbol{W} \hat{h^{\prime}_{i}}
\end{equation}
Using the softmax activation function, the attention coefficients are normalized across neighbors and multiplied by the adjacency matrix $A_{ij}$, which gives higher importance to node pairs closer in distance, reflecting the physical principle that the strength of an intermolecular bond decreases as the bond distance increases. The normalized attention coefficient $a_{ij}$ is given by:
\begin{equation}
    a_{ij} = softmax( e_{ij})*A_{ij}
\end{equation}
\noindent Each feature is then updated as a linear combination of neighboring node features $\hat{h^{\prime\prime}_i}$.
Finally, the gated attention mechanism is employed to give the transformed set of node features: $\hat{h^{\prime}_i}$:
\begin{equation}
    \hat{h^{\prime}_i} = \phi (\mathrm{\boldsymbol{U}}(\hat{h_i} || \hat{h^{\prime\prime}_i}) + b) \hat{h_{i}} + (1 - \phi (\mathrm{\boldsymbol{U}}(\hat{h_i} || \hat{h^{\prime\prime}_i}) + b)) \hat{h^{\prime\prime}_i}
\end{equation}
\noindent where \textbf{U} $\mathbb{\in R}^{2F \times 1}$ is a learned vector, $b$ is a learned scalar, and $\phi$ is the activation function. These features are passed through a multilayer perceptron (MLP). For binary classification of the activity of the protein-ligand complex, the sigmoid activation function is applied, and the binary cross-entropy loss function is used. For regression, the ReLU activation function is applied, and the mean squared error loss function is used.

The distinct featurization of our GNNs reduces the feature size of the protein, which can contain a large number of atoms, and enhances the feature size of the ligand, which is typically a small molecule. This shift reduces redundancy in the protein representation and focuses computational resources on the improved atomic representation of the ligand. While the Graph-CNN developed by Torng et al. also involved a large number of ligand features, more than half were at the bond level \cite{torng2019graph}. Because our GNNs include only atom-level features, we examined the physical properties behind the chosen bond-level features and chose atom-level features that impart the same physical information. For example, Torng et al.~included an encoding for whether a bond was in a ring; similarly, we included an encoding for whether an atom was in a ring. Torng et al.~also had three different features corresponding to bond type (single bond, double bond, and triple bond); we consolidated these features into a single atom-level feature, hybridization, which describes the bonding properties of an atom.

\begin{table*}
  \caption{Features associated with each atom in the protein and ligand for the Graph-CNN model described by Torng et al. \cite{torng2019graph}, the GNN described by Lim et al. \cite{lim2019predicting}, and GNN\textsubscript{F} and GNN\textsubscript{P} described in this contribution. Features are associated with the atom unless otherwise noted.}
  \label{tab:features}
  \begin{tabular}{lcccccc}
\hline
 & \multicolumn{2}{c}{Graph-CNN} & \multicolumn{2}{c}{GNN} & \multicolumn{2}{c}{GNN\textsubscript{F} \& GNN\textsubscript{P}} \\
     & \multicolumn{2}{c}{Torng and Altman\cite{torng2019graph}} & \multicolumn{2}{c}{Lim et al. \cite{lim2019predicting}} &  \multicolumn{2}{c}{(Current Work)} \\
\hline
Feature   &   Protein  &   Ligand  &   Protein  &   Ligand  &   Protein  &   Ligand \\
\hline
Atom Type     & x           & x         & x            & x          & x           & x   \\
Atom Degree   & x           & x         & x            & x          &             & x   \\
\textit{N} Hydrogen Atoms   & x         & x            & x          & x     &     & x   \\
Implicit Valence  & x       & x         & x            & x          &             & x   \\
Aromaticity   & x           & x         & x            & x          &             & x   \\
Atom in Ring  &             &           &              &            &             & x   \\
Residue Type  &             &           &              &            & x           &     \\
Hybridization &             &           &              &            &             & x   \\
Formal Charge &             &           &              &            &             & x   \\
Single Bond\textsuperscript{\emph{a}}  &            &  x         &              &            &             & x\textsuperscript{\emph{b}}    \\
Double Bond\textsuperscript{\emph{a}}  &            &  x         &              &            &             & x\textsuperscript{\emph{b}}    \\
Triple Bond\textsuperscript{\emph{a}}  &            &  x         &              &            &             & x\textsuperscript{\emph{b}}    \\
Bond Aromaticity\textsuperscript{\emph{a}}  &       &  x         &              &            &             & x\textsuperscript{\emph{b}}    \\
Conjugation\textsuperscript{\emph{a}}  &            &  x         &              &            &             & x\textsuperscript{\emph{b}}    \\
Bond in Ring\textsuperscript{\emph{a}} &            &  x         &              &            &             & x\textsuperscript{\emph{b}}    \\
\hline
  \end{tabular}
  
  \textsuperscript{\emph{a}} Bond feature;
  \textsuperscript{\emph{b}} bond feature is indirectly considered by a corresponding atom-level feature that captures the same physical property.
\end{table*}

\subsection{Dataset Preparation}

\subsubsection{Classification Model Datasets}

Machine learning models for training PLI prediction require data on target proteins, ligands/compounds, and interactions between them.  In this work, our goal is  to improve the degree to which the graph-based model can be generalized while also maintaining accuracy. We accomplish this by enlarging the dataset used to train our model and including a variety of targets. We collected and curated protein-ligand complexes from two public datasets, DUD-E \cite{mysinger2012directory} and PDBbind \cite{wang2004PDBbind, wang2005PDBbind}, which are described in further detail below. The datasets consist of protein-ligand complexes and their docking affinities, while some samples include experimentally derived binding affinities. Prediction is based on accessing ligands as active (active-interact molecules or positive) or inactive (set of decoys or negative) depending on whether the ligand is able to bind with the protein, as described in more detail below. Table \ref{tab:tab2} shows counts of the targets, ligands, and their various complexes for both datasets. 
\begin{table}[H]
\caption{Number of active complexes, inactive complexes, and protein targets in the PDBbind, DUD-E, IBS, and SARS-CoV-2 datasets used to create the training and test sets in this work.}
\begin{center}
\begin{sc}
\begin{tabular}{lccc}
\hline
Dataset & Targets  & Total Actives & Total In-Actives\\
\hline
PDBbind2018 & 991 & 1,418 & 4,804  \\ 
DUD-E & 96  & 46,145 & 16,996,568 \\
SARS-CoV-2 & 7 & 56,191 & 56,191 \\
\hline
\end{tabular}
\label{tab:tab2}
\end{sc}
\end{center}
\end{table}
To confirm that these two datasets offer diversity in terms of both target protein and different functionality of the ligand, we computed pair similarities between the targets and ligands, which can be found in the supporting information Figure S\ref{fig:similarity_SI}. Target similarities were determined by computing the homology between each pair, while ligand similarities were taken as the Dice similarity coefficient of the Morgan fingerprints (diameter = 6) of each pair. In both datasets, the target similarity is centered around 40\%, with DUD-E having a more diverse target set than PDBbind. The opposite is true for ligand similarity. While both datasets have mainly dissimilar ligands, the PDBbind dataset offers more diversity in ligand structure. Combining these datasets for training leads to a highly diverse training set in terms of both target proteins and ligand molecules.

\noindent \textbf{DUD-E}: The DUD-E dataset consists of pairs of experimentally verified active complexes and property-matched inactive pairs, called decoys \cite{mysinger2012directory}. 
The dataset was originally designed to test benchmark molecular docking programs by providing challenging decoys, but some have noted that the dataset suffers from limited chemical space and biases \cite{smusz2013influence, chen2019hidden}. In their analysis of the DUD-E dataset, Chen et al. generated a docked subset \cite{chen2019hidden}, which we use here. 
Ligands from the experimentally verified ChEMBL dataset were designated as positive, while the generated decoys and their docked structures were considered negative. These files were parsed into individual standard database formats (SDF) files for each ligand and corresponding docking pose.

Ligands with a molecular weight above 500 Da were removed, and the docked structures were converted to a machine-readable format using RDKit \cite{Landrum2014}. The target proteins were pulled from the original DUD-E set, cleaned of water molecules, and matched with their designated ligands. Targets were screened with a homology test to understand the diversity of targets present in the dataset, and none had a similarity greater than 90\%. Each complex was processed to crop the target protein and retain atoms within a distance of 8 Å from the ligand. Because the DUD-E dataset is biased towards inactive complexes due to the large number of decoys, equal counts of active and inactive samples were collected in regards of a target with no consideration of a specific complex. A complete 1:1 active-to-inactive ratio was not always achieved; however, the imbalance was found to be minimal and to have no affect on performance. Samples from the experimentally verified ChEMBL dataset were labeled as active, while decoy samples were denoted with the key word ZINC.
\label{dude}

\noindent \textbf{PDBbind}: The PDBbind dataset contains experimentally verified protein-ligand complexes from the Protein Data Bank \cite{wang2004PDBbind, wang2005PDBbind}.
Binding poses for a refined set of the protein-ligand complexes were generated by docking calculations \cite{GitHubRM95:online}.  
A 90\% homology test was run on this set, which resulted in a small number of proteins being removed because of a high level of similarity. As with the DUD-E dataset, the docked structures were converted to a machine-readable format using RDKit. The root-mean-square distance (RMSD) was used to label ligands as positive if they maintained an RMSD less than or equal to 2\AA  compared to the original crystal structure, and negative if the RMSD was greater than or equal to 4\AA. Molecules with RMSDs between these thresholds were removed. The viable molecules and their target proteins were processed into a dataset of protein atoms cropped within a distance of 8\AA. As PDBbind data were significantly limited, all available samples were used. This resulted in a greater inclusion of negative samples than positive samples. 
\label{PDBbind}

\noindent \textbf{SARS-CoV-2 dataset }: The SARS-CoV-2 dataset is composed of seven protein targets: M\textsuperscript{Pro}\textunderscore6WQF, NSP15\textunderscore6XDH, M\textsuperscript{Pro}\textunderscore6LU7, PL\textsuperscript{Pro}\textunderscore6W9C,PL\textsuperscript{Pro}\textunderscore6WRH, ADRPNSP3\textunderscore6W02, and NSP10-16\textunderscore6W61 with three main protease (M\textsuperscript{Pro}, two papain-like cysteine protease, one open reading frame, and three non-structural proteins. Large ligand libraries composed from FDA \cite{FDA} and manually curated antiviral data were used to generate the docked complexes with each target.  
Ten docking poses were calculated for each complex with the qvina docking program\cite{trott_olson_2009} through a custom non-covalent pipeline. The docking data was parsed and cropped to the 8 Å threshold in the same manner as the DUD-E and PBBind datasets. Similar to PDBbind, positive and negative samples are determined with RMSD. Protease data were largely directed into the training set while the other targets, with the exception of on non-structural protein, were directed to the test set. This variety allows the inclusion of the critical targets of SARS-CoV-2 viral life cycle in both the training and the test sets.

\noindent \textbf{IBS dataset}: The IBS dataset is created using 486,232 synthetic compounds from the IB screening database (www.ibscreen.com). 
All ligands were fed as an input to the our GNN models paired against M\textsuperscript{Pro} and NSP15 target. 
Because the IBS dataset only contains ligands and we did not dock the IBS molecules with their 
corresponding receptors, we used our GNN\textsubscript{P} model to evaluate these complexes. We chose 
SARS-CoV-2 (M\textsuperscript{Pro}) and SARS-CoV-2 non-structural protein endoribonuclease 
(NSP15) as target proteins to study the performance of our model. These are exactly the same targets we 
have used for our creating our SARS-CoV-2 dataset.  Our team has been actively workin1g on the 
development of covalent electrophiles and non-covalent inhibitor candidates against the viral 
proteases, this gave us ready access to the protein pockets and active compounds that were binding with 
these targets. In addition, we performed homology tests on M\textsuperscript{Pro} and NSP15 against the 
targets in the DUD-E and PDBbind training sets to quantify the similarity of these new targets to our 
larger dataset. M\textsuperscript{Pro} showed an average similarity of 40\% with the DUD-E training 
targets and 48\% with the PDBbind training targets. NSP15 showed an average similarity of 
55.72\%  with DUD-E targets and 52.26\%  with PDBbind targets.

\subsubsection{Regression Model Datasets}
In this section, we discuss the datasets used for the regression models noting that part of these datasets are created from the same sources used for the classification models. Our regression models are composed of the same attention heads as in the classification models and, therefore, include the same domain-level information from the protein and ligand atoms. The main difference in our classification and regression models is the final activation layer, as shown in Figure \ref{fig:GNNSP}. We perform two regression experiments referred as
Experimental Binding Affinity (EBA)and pIC\textsubscript{50} prediction.  It is important to highlight that some of these properties such as  pIC\textsubscript{50} cannot be accurately modeled through physics based modeling methods. Table \ref{tab:EBA_dataset} and \ref{tab:pic50_dataset} gives a summary of the target and ligands distribution in various regression datasets. 

For Experimental Affinity experiments, we consider three data sources: 1) PDBbind2016, 2) PDBbind2018, and 3) PDBbind2019 \cite{wang2005PDBbind}. We consider just crystal poses from the PDBbind2016 general, refined, and core datasets. For the PDBbind2018 dataset, we used both the docked and crystal structures as two independent datasets for two independent experiments. For the docked dataset, we used the same targets and split as used for classification experiments. For the crystal-only dataset, we used all the available targets without any homology-similarity screening.  For the PDBbind2019 dataset, we considered just a handful of crystal structures as a part of independent test set. The PDBbind2019 dataset is a structure-based evaluation set comprises targets that have been added to PDBbind2019 after the year 2016 and thus are not a part of PDBbind2016 and have a GDC similarity less than 65\% to the training targets.

\begin{table}[ht]
\begin{center}
\begin{small}

\resizebox{\textwidth}{!}{\begin{tabular}{lccccccc}
\hline

Dataset & Total  & Total   & Train   & Test  & Train  & Test  & Total  \\
 &  targets &  ligands  &  targets  &  targets & samples &  samples &  samples \\
\hline
PDBbind2018-EBA (with docked poses) & 1,278 & 1278 & 1,023 & 255 & 6,485 & 1557 & 8042\\
PDBbind2018-EBA (crystal only) & 10,375 & 10,375 & 8,300 & 2,075 &  8,300 & 2,075 & 10,375\\
PDBbind2016-EBA (crystal only) & 11,674 & 11,674 & \textsuperscript{\emph{a}} & \textsuperscript{\emph{a}} &  \textsuperscript{\emph{a}} & \textsuperscript{\emph{a}} & 11,654\\
PDBbind2019-EBA (crystal only) & 190 & 190 & - & - &  - & - & 190\\
\hline
\end{tabular}}
\textsuperscript{\emph{a}} Refer to 
Table S\ref{tab:table_PDBbind2016_splits} in the supporting information for detailed dataset splits.
\end{small}

\end{center}

\caption{Summary of data used for training and testing EBA regression models.} 
\label{tab:EBA_dataset}
\end{table}

For the PDBbind2016 dataset, we prepared various train-test splits, which are summarized in Table S\ref{tab:table_PDBbind2016_splits}. 
The experimental binding affinity is estimated in terms of K\textsubscript{i} and K\textsubscript{d}, which refer to the inhibition and dissociation constants, respectively. These properties collectively determine the binding affinity of a molecule towards a receptor. The PDBbind repository provides a database of protein-ligand complexes along with their experimentally measured data. Here, we define the experimental binding affinity as  $-log(\frac{K_{i}}{K_{d}})$, which is used as the label in the regression model. All decoys from the PDBbind2018 dataset were labeled with the experimental affinity of the corresponding crystal structure.

We focus on the pIC\textsubscript{50}(the inverse log of the half maxi-mal inhibitory concentration, IC\textsubscript{50}), which is an experimentally measured property that captures the potency of a therapeutic candidate towards a protein target where higher values indicate exponentially more  potent inhibitors.  For the pIC\textsubscript{50} data, we used a combination of the DUD-E and PDBbind2016 datasets. From DUD-E, we included active ligands for 65 of the DUD-E targets, as only active protein-ligand pairs have an associated experimental affinity in the ChEMBL repository. We considered the top-scoring docked pose for each protein-ligand complex in the DUD-E dataset because there were no crystal structures available.  

\begin{table}[h]
\begin{center}
\begin{tabular}{lccc}
\hline

Dataset & Total   & Train   & Test \\
 &  samples &  samples  &  samples \\
\hline
PDBbind2016-pIC\textsubscript{50} & 4,576 & 3676 & 900 \\
DUD-E-pIC\textsubscript{50}-only-top-docked-pose & 3,706 & 3,105 & 601 \\

\hline
\end{tabular}

\end{center}

\caption{Table summarizing the dataset used for pIC\textsubscript{50} regression models.}
\label{tab:pic50_dataset}
\end{table}
For the PDBbind2016 data, all crystal poses with an experimentally measured IC\textsubscript{50} were used. We retained the same 80:20 split for the PDBbind2016 dataset as used for the experimental affinity dataset. Table \ref{tab:pic50_dataset} gives details of the dataset used for pIC\textsubscript{50} regression.

We also considered an independent dataset associated with SARS-CoV-2 targets. To prepare this dataset, we used crystal structures of SARS-CoV-2 main protease (M\textsuperscript{Pro}) bound to non-covalent inhibitors that have an associated experimental IC\textsubscript{50}. The complex 7LTJ \cite{7ltj} is obtained as a result of non-covalent inhibition of MCULE-5948770040 compound with M\textsuperscript{Pro} (PDB-ID: 7JUN) discovered using our previous high throughput virtual screening as a part of the U.S. Department of Energy National Virtual Biotechnology Laboratory (NVBL) project \cite{clyde2021high}. When a protein-ligand complex had multiple measured IC\textsubscript{50}, we used an average of the values as the label. The PDB-IDs of the other targets with the constituting protein and ligand are listed in Table \ref{tab:sars_cov_pdb}.

\begin{table}[h]
\begin{center}
\begin{tabular}{lccc}
\hline

PDB-ID & Target & Inhibitor \\
\hline
7LTJ \cite{7ltj} & M\textsuperscript{Pro} & MCULE-5948770040 \\
7L0D \cite{7l0d}& M\textsuperscript{Pro} & ML188 \\
7LME \cite{7lme}& M\textsuperscript{Pro} & ML300 \\    
7L11 \cite{zhang2021potent} & M\textsuperscript{Pro} & Compound 5\textsuperscript{\emph{a}}\\
7L12 \cite{zhang2021potent} & M\textsuperscript{Pro} &Compound 14\textsuperscript{\emph{a}}\\
7L13 \cite{zhang2021potent} & M\textsuperscript{Pro} & Compound 21\textsuperscript{\emph{a}}\\
7L14 \cite{zhang2021potent} & M\textsuperscript{Pro} & Compound 26\textsuperscript{\emph{a}}\\
\hline
\end{tabular}

\end{center}

\caption{SARS-CoV-2 targets with experimentally assessed non-covalent inhibitors used for testing pIC\textsubscript{50} models.}
\label{tab:sars_cov_pdb}

  \textsuperscript{\emph{a}} As defined by Zhang et al.\cite{zhang2021potent}.

\end{table}
\vskip 0.2in
\label{subsection:regression dataset}
\subsection{Hyperparameter Optimization}

Hyperparameters such as network depth, layer dimension, and learning rate can have a large effect on model training and the weights in the final realized model. Therefore, we performed a number of trainings to examine combinations of learning rate, number of attention heads, and layer dimension. 
The hyperparameters that performed best were a learning rate of 0.0001, two attention heads, and a dimension of 70. These parameters resulted in an average test AUROC of 0.864. The combinations of these parameters are summarized in Table S\ref{tab:hypTable_SI}, and all successfully completed hyperparameter experiments can be found in Table S\ref{tab:hypTableSI} along with further explanation of the trials.

\section{Results and Discussion}
\subsection{GNN Classification Models}

Our primary goal in this work is to develop a model that has high accuracy and that can be generally applied for predicting PLI and activity using distinct atom- and bond-level features (domain awareness) for the protein and ligand. To this end, we would like to understand the effects of the number of protein targets and the number of protein-ligand complexes per target; therefore, we train the GNN\textsubscript{F} and GNN\textsubscript{P} models on a variety of datasets. The datasets consisted of either 17, 79, or 96 targets from the DUD-E dataset and all available data from the PDBbind dataset, which are summarized in Figure S\ref{fig:comparison} in the SI. In all cases, the accuracy greatly increases from 0.723 to 0.879 when the number of targets is increased from 17 to 79, and then decreases to 0.842 when the number of targets is further increased to 96. The accuracy of each experiment is shown in Table S\ref{tab:ExprAccSI}. In each training set, we examined the effects of having 1,000 complexes per target or 2,000 complexes per target. When using 17 DUD-E targets, including more complexes per target did not improve the accuracy, while in both the 79 and 96 DUD-E target sets, the accuracy was improved to over 90\% when the dataset consisted of 2,000 complexes per target. Notably, the improvement was greater for the 96 DUD-E target set. It is important to note that all training sets had an equal distribution of positive and negative samples and complexes were randomly divided into training and test sets with an 80:20 split. 

In addition to making sure the model can be generally applied, we are interested in developing a model that does not require the docked structure to be known before inferences can be made. We performed the same experiment using varying numbers of DUD-E targets and complexes on our GNN\textsubscript{P} model, which does not require advance knowledge of the protein-ligand interaction to predict activity. The same overall trends were observed for GNN\textsubscript{P} as for GNN\textsubscript{F} but with reduced accuracy, as shown in Table S\ref{tab:ExprAccSI}. The highest scoring GNN\textsubscript{P} model was that trained on 79 DUD-E targets with 2,000 complexes per target with an accuracy of 0.880\%, which is only 3.2\% lower than the GNN\textsubscript{F} model trained on the same dataset. Though decreased accuracy was observed with GNN\textsubscript{P}, its advantage relies on knowledge of only the separated protein and ligand structures, greatly reducing required preprocessing steps and increasing the throughput of the trained model.

\begin{table}
  \caption{Comparison of test dataset results for our GNN\textsubscript{F} and GNN\textsubscript{P} models on the various test sets described in the text, along with representative examples from the literature. The top scores are shown in bold.}
  \label{tab:data}
  \begin{tabular}{lcccc}
\hline
Method & Test Accuracy & Sensitivity & Specificity & Reference \\
\hline
GNN\textsubscript{F} Overlap & \textbf{0.979} & 0.840 & \textbf{0.970} & Current Work \\
GNN\textsubscript{P} Overlap & 0.958 & \textbf{0.870} & 0.910 & Current Work \\
GNN\textsubscript{P} Distinct & 0.855 & 0.590 & 0.910 & Current Work \\
GNN\textsubscript{F} Distinct & 0.951 & 0.690 & \textbf{0.970} & Current Work \\
GNN\textsubscript{F} Base & 0.934 & 0.660 & \textbf{0.970} & Current Work \\
GNN\textsubscript{P} Base & 0.845 & 0.580 & 0.900 & Current Work \\
Docking  & 0.591 &  &  & Current Work \\
GNN & 0.968 & 0.830 & \textbf{0.970} &  Lim et al.\cite{lim2019predicting} \\
CNN  & 0.904 &  &  & Gonczarek et al.\cite{gonczarek2016learning} \\
CNN  & 0.868 &  &  & Ragoza et al.\cite{ragoza2017protein} \\
CNN  & 0.855 &  &  & Wallach et al.\cite{wallach2015atomnet}\\
Graph-CNN  & 0.886 &  &  & Torng and Altman\cite{torng2019graph} \\
\hline
  \end{tabular}
\end{table}

To assess the ability of our GNN models to be generally applied, we produced three test sets of varying similarity to the training set (Table \ref{tab:data}).
Our base dataset consists of 79 targets from DUD-E and 991 targets from PDBbind randomly distributed 
into the training and test sets, with roughly 2,000 samples per target and equal distribution of positive 
and negative samples. We also considered a dataset with overlapping targets but with novel complexes 
distributed into the training and test sets. Approximately 2,000 additional samples that were withheld 
from the initial training were collected for each training target to show the effect of target overlap 
between training and test samples. Furthermore, we created a distinct dataset comprised of samples not 
used for training for the same target distribution in the base set. Table \ref{tab:data} shows the 
results of these tests in terms of test set accuracy, sensitivity, and specificity, along with some 
representative examples from the literature. The best performance is attained for the 
GNN\textsubscript{F} model for the overlapping target dataset with an accuracy of 0.979, followed by the 
GNN\textsubscript{P} model on the same dataset with an accuracy of 0.958. Additionally, the models show 
improved performance on the distinct sample set as compared with the base sample set. The accuracy 
increases from 0.934 to 0.951 for the GNN\textsubscript{F} model and from 0.845 to 0.855 for the 
GNN\textsubscript{P} model. The close similarity of the base, distinct, and overlap test set accuracies 
of the GNN\textsubscript{F} model indicates  that this  model could be generally applied. 
GNN\textsubscript{P}, however, showed  reduced accuracy for the base and distinct test sets as compared 
with the overlap test set, indicating decreased capability to be generally applied. Overall, each 
implementation shows significantly improved performance in terms of prediction compared to docking. We 
can see that the overlap set produces a slight increase in
AUROC, decrease in specificity, and 
significant increase in sensitivity.
\subsubsection{Top-N ranks}
\begin{figure}[H]
\begin{center}
    \centerline{
        \includegraphics[width=0.5\textwidth]{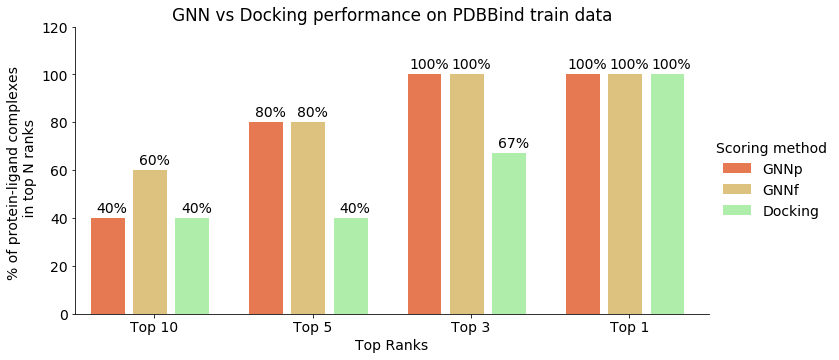}
    \includegraphics[width=0.5\textwidth]{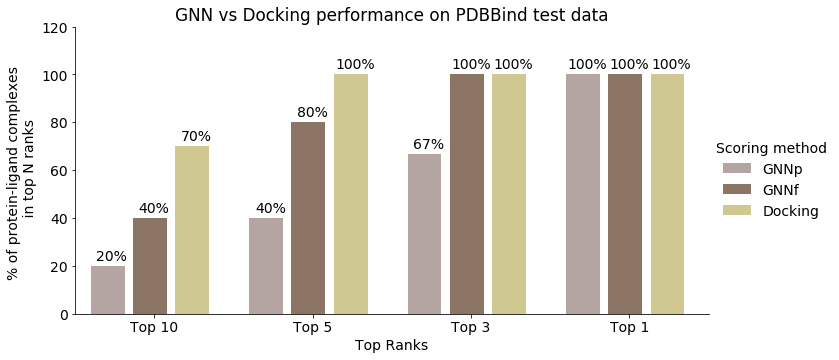}}
    \caption{Comparison of the GNN\textsubscript{P} and GNN\textsubscript{F} models with docking. Each bar corresponds to the percentage of protein-ligand complexes identified in top-N ranks which have an RMSD less than 2 \AA~from the crystal structure. }
    \label{fig:top_n_ranks_m}
\end{center}
\end{figure}

We then assessed the model's ability to identify top-scoring 3D poses in each protein-ligand complex 
from our PDBbind repository described in the dataset preparation section. The 'best' docked pose is quantified as having an RMSD of less than 2 \AA~with respect to the crystal structure. In this analysis, we 
measure not only the ability of the model to identify an active protein-ligand pose but also its ability to 
identify the best pose among multiple docked poses. Figure \ref{fig:top_n_ranks_m} shows the percent of 
complexes found in the top-N ranks. 
In each rank, both the GNN\textsubscript{F} and GNN\textsubscript{P} models outperformed or matched the performance of docking when using complexes from the training data. However, on the test data, docking showed better performance for certain ranks, while both docking and GNN\textsubscript{F} identified 100\% of the protein-ligand complexes. For the top rank, all three methods showed 100\% identification. If we consider the percent targets with at least one pose in the top-N ranks, all three models show equivalent performance (see Fig. S\ref{fig:top_n_ranks_docking}).

In addition, we tested our models on datasets that include molecules from ChEMBL for DUD-E targets and SARS-CoV-2-target-specific data. Additional ChEMBL data were collected for a small subset of DUD-E targets extracted from the initial test set and implemented without docking for the GNN\textsubscript{P} model. These molecules were prepared and matched with a pocket from the designated target. Roughly 2,000 samples were prepared for two SARS-CoV-2 targets, M\textsuperscript{Pro} and NSP15, in a relatively balanced split of positive and negative samples. The GNN\textsubscript{F} and GNN\textsubscript{P} models trained on DUD-E and PDBbind then were  used for inference on this set. Among all the experiments, GNN\textsubscript{P} performed best  with the ChEMBL data, showing an ROC of 0.596. 
The models performed relatively low on the SARS-CoV-2 targets, showing very low ROCs of 0.415 and 0.281 for the GNN\textsubscript{P} and GNN\textsubscript{F} models, respectively. This can be attributed to the fact that the SARS-CoV-2 target (M\textsuperscript{Pro}) used in this dataset has an average similarity of 40\% with DUD-E training targets and 48\% similarity to the PDBbind training data and therefore represents extrapolation rather than interpolation. Notably, graph neural networks are known to perform poorly on nonlinear extrapolation tasks far from the training data \cite{xu2020neural}. 

\begin{figure}[H]

\begin{center}
\centerline{
\includegraphics[scale=0.30]{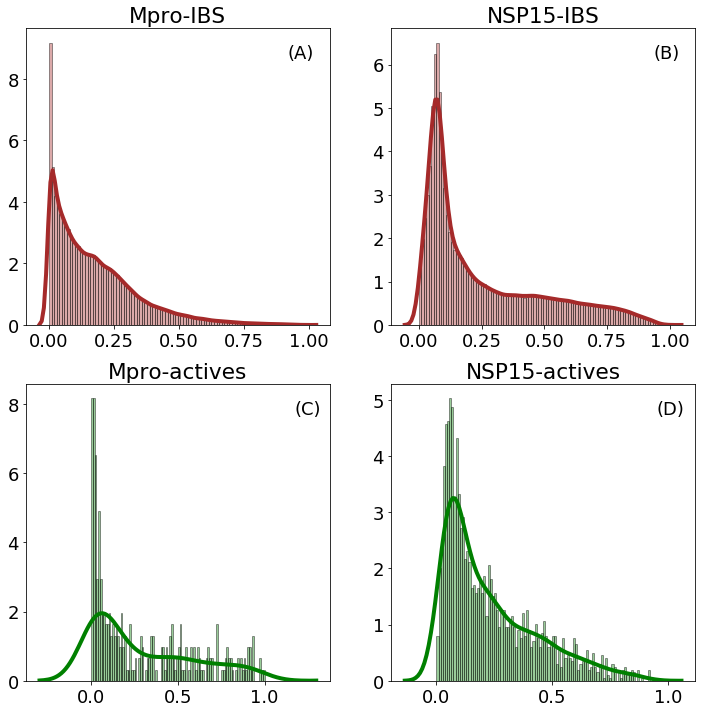}}
    \caption{Binding probability distribution for IBS molecules with M\textsuperscript{Pro} and NSP15 as targets. Figures (A) and (B) correspond to the predicted binding probability for NSP15 and M\textsuperscript{Pro} targets against IBS molecules. Figures (C) and (D) correspond to the predicted binding probability on active molecules for M\textsuperscript{Pro} and NSP15 respectively (for each plot, the x-axis denotes the predicted probability and y-axis denotes the density of molecules).}
    \label{fig:fig6}
\end{center}
\end{figure}

We next investigated the performance of the GNN\textsubscript{P} model  on IBS molecules with M\textsuperscript{Pro} and NSP15 protein targets. Their binding probability distributions are shown Figure \ref{fig:fig6}. The predicted activity distributions for NSP15 actives and IBS molecules are similar, suggesting that GNN\textsubscript{P} can identify a majority of compounds from IBS with potential to bind to the NSP15 receptor. The predicted probability distribution of M\textsuperscript{Pro} target actives shows that GNN\textsubscript{P} can identify a high percentage of molecules with potential binding affinity against the receptor, though it struggled to identify many of the IBS molecules. Theoretically, compounds with lower molecular weight (smaller size) possess a greater tendency to bind against a target. Most of the IBS compounds have a molecular weight between 250 and 700 Dalton (see Fig. S\ref{fig:IBS_mol_weight}). For the M\textsuperscript{Pro} target, our results indicate that the binding probability is centered near zero and the majority binding probability is  under 0.5. For the NSP15 target, the binding probability is centered closer to 0.1 with a small number of compounds having probabilities greater than 0.5. We observe that GNN\textsubscript{P} consistently performed better in relation to the NSP15 target. Contributing factors to this increased performance include a larger number of available active compound samples for the NSP15 targets; that is, 2,157 samples as opposed to 307 samples available for the M\textsuperscript{Pro} target. The NSP15 target has a disproportionately larger binding pocket than that of M\textsuperscript{Pro}, which also attributes to the improved performance.
\subsection{GNN as a Regression Model}

Predicting the binding affinity of a protein-ligand complex plays a critical role in identifying a lead molecule that binds with the protein target; however, the experimental measurement of protein-ligand binding affinity is laborious and time-consuming, which is one of the greatest bottlenecks in drug discovery. On the other hand, half maximal inhibitory concentration (IC\textsubscript{50}) provides a quantitative measure of the potency of a candidate to inhibit a  protein target and is typically estimated from experiments. If we can predict affinity and potency of a specific ligand to a target protein quickly and predict their interactions accurately, the efficiency of in silico drug discovery would be significantly improved. To this end, we modified both GNN models to perform regression in order to predict the biophysical and biochemical properties (affinity and potency) of protein-ligand complexes. Our regression models are composed of the same attention heads as in the classification models and, therefore, include the same domain-level information for the protein and ligand atoms. We evaluated these models on each of the datasets discussed in the Methods section.

 \noindent \textbf{Experimental Binding Affinity (EBA) regression experiments}:\\
To predict binding affinity and assess the performance of our model, we performed three experiments using three different datasets: 
1) PDBbind2018-docked, 2) PDBbind2018 crystal, and 3) PDBbind2016 crystal structure dataset. 
Since most of the deep learning models that we compare in this section are trained on complexes from the PDBbind2016 database, training our models on the same dataset helps to obtain a comparison with the previous models \cite{fast, stepniewska2018development, jimenez2018k}. We also tested our models on the PDBbind2016 core set, which is a refined subset filtered on the basis of protein-sequence similarity. 

First, we assessed the performance of the GNN models trained using PDBbind2018 data docked and crystal-only data as shown in Table \ref{tab:8}. Notably, the addition of more unique protein-ligand complexes improves the performance, while the addition of multiple docking poses for fewer complexes decreases performance, as the Pearson and Spearman correlations are low on the docked dataset as compared to the crystal-only dataset. We compared our methods with Pafnucy \cite{stepniewska2018development}, which is a 3D CNN for protein-ligand affinity prediction that combines the 3D voxelization with atom-level features. Among all the results, the best performance is achieved by the GNN\textsubscript{P}-EBA model, and overall, both GNN\textsubscript{P} and GNN\textsubscript{F} outperform Pafnucy on both the docking and crystal-only datasets. 

\begin{table}[H]
    \centering
    \begin{tabular}{cccccc}
    \hline
    MODEL & RMSE & MAE & Pearson r & Spearman r & $r^2$\\
    \hline
    GNN\textsubscript{F}-EBA-docking &  1.74 & 1.41 & 0.40 & 0.39 & 0.16 \\
    GNN\textsubscript{P}-EBA-docking & 1.61 & 1.32 & 0.49 & 0.49 & 0.24 \\
    GNN\textsubscript{F}-EBA-crystal-only & 1.68 & 1.32 & 0.45 & 0.46 & 0.21 \\
    GNN\textsubscript{P}-EBA-crystal-only &  \textbf{1.48} & \textbf{1.16} & \textbf{0.59} & \textbf{0.59} & \textbf{0.35} \\
    Pafnucy \cite{stepniewska2018development} (docking)$^a$ & 1.6 & 1.34 & 0.52 & 0.50 & 0.27 \\
    Pafnucy \cite{stepniewska2018development} (crystal-only)$^b$ & 1.86 & 1.50 & 0.38 & 0.37 & 0.14 \\
\hline
    \end{tabular}
    \caption{Performance comparison of our GNN models in predicting experimental affinity on the PDBbind dataset. The top scores are shown in bold.}
    \label{tab:8}

  \textsuperscript{\emph{a}} 255 test targets from our PDBbind2018-EBA-docking dataset using all docked poses for evaluation;
  \textsuperscript{\emph{b}} PDBBind2018-EBA-crystal-only test dataset. 

\end{table}

In addition, for non-docking experiments, we used the PDBbind2016 general and refined datasets for training 
and the PDBbind2016 core set for testing. We assessed the performance of our methods for predicting the affinity through comparison with the K\textsubscript{DEEP} \cite{jimenez2018k} and FAST \cite{fast} models. Our analysis on the PDBbind2016 core set shows that the GNN\textsubscript{P}-EBA model performs similarly to the FAST model \cite{fast}, while K\textsubscript{DEEP} shows the highest performance (see Table S\ref{tab:PDBbind_2016_core_set} for detailed comparison). We also report the performance of our model on the dataset used for training and evaluating the FAST model (see Table S\ref{tab:FAST_GNN_same_split} for details) using the same train-validation-test splits as provided by the authors over the PDBbind2016 general (G) and refined (R) sets \cite{fast}.  Out of all the combinations of the general and refined sets, we achieved the best performance with the GNN\textsubscript{P} model trained on the general set.

Our results on the PDBbind2018 and PDBbind2016 datasets suggests that with our proposed GNN frameworks, we achieve top performance compared with prior deep learning methods for binding affinity prediction while preserving the spatial orientation of the protein and ligand. While K\textsubscript{DEEP} is a purely 3D CNN-based network for protein-ligand affinity predictions, the FAST model uses a graph-based network, but utilizes the 2D graph representation which does not include the non-covalent interactions. This structural information is critical for understanding PLI and its impact on estimating affinity. To tackle these issues, we devised our graph-based models to specifically include all the atom- and bond-level information while using the 3D structures of the protein and ligand. The graph representation not only encloses atom- and bond-level information, but also retains the spatial information associated with the protein-ligand complexes, thus enabling us to include all necessary information associated with the natural binding state of the protein and ligand. Our GNN\textsubscript{F} model accounts for intermolecular interactions between the protein and ligand, which are not captured in the FAST model.

To investigate the generalizability of our GNN models in predicting the binding affinity of unseen and novel targets, we compare the performance of our GNN and various models on the PDBbind2019 structure-based evaluation dataset (Table \ref{tab:PDBbind_2019_sbed}). The PDBbind2019 structure-based evaluation dataset is composed of targets that are novel from the PDBbind2016 in terms of their addition to the database as well as sequence similarity. The prediction results of the GNN models are better than those of previous models, as shown in Table \ref{tab:PDBbind_2019_sbed}, and most importantly, our GNN\textsubscript{F}-EBA model performs as accurate as  Pafnucy. Our results demonstrate that even with limited 3D structural data in terms of the size typically needed to train deep learning models, we achieved relatively more accurate generalization with the GNN model. In addition, our GNN outperforms FAST and K\textsubscript{DEEP} in terms of generalizability given the scarcity of available structural data.

\begin{table}[h!]
    \centering
    \begin{tabular}{ccccc}
    \hline
    MODEL & RMSE & MAE & Pearson r & Spearman r \\
    \hline
    GNN\textsubscript{F}-EBA & 1.39 & \textbf{1.10} & 0.49 & 0.50  \\
    GNN\textsubscript{P}-EBA & 1.52 & 1.22 & 0.42 & 0.46  \\
    Pafnucy \cite{stepniewska2018development} & \textbf{1.38} & 1.11 & \textbf{0.52} & \textbf{0.52}\\
    FAST \cite{fast} & 1.48 & 1.21 & 0.42 & 0.40 \\
    K\textsubscript{DEEP} \cite{jimenez2018k} & 1.42 & 1.13 & 0.48 & 0.47 \\
    
    \hline
    \end{tabular}
    \caption{Performance comparison of our GNN models in predicting experimental affinity on the PDBbind2019 structure-based evaluation dataset. The results for the FAST method are reported for its 3D CNN model. The top scores are shown in bold.}
    \label{tab:PDBbind_2019_sbed}
\end{table}

Finally, to expand the scope of our experiment, we also trained our model on a physics-based docking affinity score. We refer to these models as Docking Binding Affinity (DBA) models. We compare our model's performance against the physics-based docking on the two DBA datasets. Our results suggest that, with a correlation score of 0.79, the GNN\textsubscript{F} model was able to reproduce some correlation between the actual and predicted affinity to an extent (see Tables S\ref{tab:DBA_dataset} and S\ref{tab:DBA_results} for detailed dataset description and results). This indicates that, with our GNN framework, we are not only capturing the details needed for predicting the binding affinity but also achieving the capability to differentiate between distinct docked poses of a protein-ligand complex and associate it with its docking score.

\noindent \textbf{pIC\textsubscript{50} regression experiments}:\\
As a next step, we tailored our GNN models to predicting pIC\textsubscript{50}. pIC\textsubscript{50} provides a quantitative measure of the potency of a candidate to inhibit a protein target, which is typically estimated from experiments. A number of methods have been developed to approximate pIC\textsubscript{50}, so we compared our pIC\textsubscript{50} prediction with existing deep-learning methods, such as DeepAffinity \cite{karimi2019deepaffinity}, DeepDTA \cite{ozturk2018deepdta} and MONN \cite{li2020monn}. While these methods were trained purely on the protein sequence and 2D SMILES representation of the ligands, our model is novel in that it considers the 3D structures of the protein and ligand  to predict pIC\textsubscript{50}, which is key to defining the inhibition rate while identifying and optimizing hits in early-stage drug discovery. To the best of our knowledge, 3D protein-ligand complexes have not been used to predict pIC\textsubscript{50} before. 

The overall performance of the GNN model is relatively low compared to the existing methods listed in the Table \ref{tab:gnn_pic50_performance}. This could be attributed to the smaller size of the dataset containing both pIC\textsubscript{50} and the corresponding crystal structures. DeepAffinity \cite{karimi2019deepaffinity}, DeepDTA \cite{ozturk2018deepdta}, and MONN \cite{li2020monn} were trained on BindingDB data, which has nearly 10 times the amount of data than that available for our dataset. In addition, we observed improvement in the Pearson correlation coefficients in predicting pIC\textsubscript{50} from 0.45 to 0.51 when using weights from the GNN-EBA model. The improvement from the baseline to the transfer learning model suggests that our model can achieve better performance if a larger dataset is used, as the GNN-EBA models are trained on comparatively larger datasets.

\begin{table}[H]
    \centering
    \begin{tabular}{cccc}
    \hline
    Model & RMSE & Pearson r \\
    \hline
    GNN\textsubscript{P}-pIC\textsubscript{50}&  1.24 & 0.45 \\
    GNN\textsubscript{F}-pIC\textsubscript{50} & 1.26  & 0.44 \\
    GNN\textsubscript{P}-pIC\textsubscript{50} best GNN\textsubscript{F}-EBA weights &  1.21 & 0.51 \\
    GNN\textsubscript{F}-pIC\textsubscript{50} best GNN\textsubscript{P}-EBA weights& 1.24 & 0.51 \\
    DeepAffinity\cite{karimi2019deepaffinity}	& \textbf{0.74}&	0.84\\
    DeepDTA\cite{ozturk2018deepdta}&	0.78&	0.85\\
    MONN\cite{li2020monn}&	0.76&	\textbf{0.86}\\
        \hline
    \end{tabular}
    \caption{Performance comparison of deep learning models in predicting pIC\textsubscript{50}. Our models were trained and tested on PDBbind2016 + DUD-E targets, whose IC\textsubscript{50} was curated from the PDBbind and ChEMBL repositories, respectively. DeepAffinity, DeepDTA, and MONN were trained and tested on BindingDB data. The top scores are given in bold.}
    \label{tab:gnn_pic50_performance}
\end{table}

To assess how critical the learned protein and ligand representations are for predicting pIC\textsubscript{50}, we predicted pIC\textsubscript{50} of a few inhibitors that were recently designed for SARS-CoV-2 Mpro, where the co-crystal structures have been solved and their IC\textsubscript{50} have been experimentally measured. From the perspective of deep learning and 3D protein-ligand complex representation, this is a much more difficult regression problem compared to the classification problem above. Our ultimate goal was to quantify the error in IC\textsubscript{50} prediction relative to the experimental measurement that could used for iterative design of potential inhibitors or lead optimization. 

\begin{table}[H]
    \centering
    \begin{tabular}{cccccc}
    \hline
    PDB-ID & 
    \shortstack{\\GNN\textsubscript{P}-pIC\textsubscript{50}} & \shortstack{\\GNN\textsubscript{P}-pIC\textsubscript{50} \\best EBA} & \shortstack{\\GNN\textsubscript{F}-pIC\textsubscript{50}} & \shortstack{\\GNN\textsubscript{F}-pIC\textsubscript{50}\\ best EBA} & \shortstack{\\Experimental \\pIC\textsubscript{50}}\\
    \hline
    7TLJ &  
    6.20 & 7.12 & 6.51 & 6.90 & 5.37 \\
    7L0D & 7.56 & 7.08 & 7.33 & 7.43 & 5.6 \\
    7LME & 7.61 & 7.07 & 7.46 & 7.59 & 5.3\\
    7L11 & 7.90 & 7.13 & 7.74 & 7.59 & 6.8\\
    7L12 & 7.96 & 7.36 & 8.26 & 7.61 & 7.74\\
    7L13 & 8.11 & 7.42 & 8.42 & 7.56 & 6.89\\
    7L14 & 8.06 & 7.16 & 7.87 & 7.67 & 6.76 \\
    \hline
    \end{tabular}
    \caption{Performance of GNN on SARS-CoV-2 M\textsuperscript{Pro} targets and some of the potential inhibitors whose IC\textsubscript{50} has been experimentally measured.}
    \label{tab:pIC50_covid_data_results}
\end{table}

Our extensive analysis on the SARS-CoV-2 M\textsuperscript{Pro} data demonstrates that both GNN models overestimate IC\textsubscript{50} by 0.92\% as compared to experimental values, as shown in Table \ref{tab:pIC50_covid_data_results}. The pIC\textsubscript{50} of our recently designed MCULE-5948770040 compound with M\textsuperscript{Pro} 7TLJ complex with GNN\textsubscript{P} model  predicted to be 6.20, which is comparable to the measured experimental value of 5.37 \cite{clyde2021high}. 
Interestingly, GNN\textsubscript{P} proved to be the best model with an average error of 0.42 Molar. It is important to highlight that GNN\textsubscript{P} gives an advantage in predicting pIC\textsubscript{50} even when the experimentally bound structure is not known. This can help rank potent candidates while screening potential libraries against a protein target or possible protein targets of a given disease, which can then be utilized  for experimental testing.

\section{Conclusion}

In this work, we devised graph-based deep learning models, GNN\textsubscript{P} and GNN\textsubscript{F}, by integrating  knowledge representation, 3D structural information and reasoning for PLI prediction through classification and regression properties of protein-ligand complexes. The parallelization of the GNN\textsubscript{P} model provides a basis for novel implementation of structural analysis that requires no docking input but instead separate protein and ligand 3D structures. The basic strategy of GNN\textsubscript{P} is to learn embedding vectors of the ligand graph and protein graph separately and combine the two embedding vectors for prediction. The featurization of GNN\textsubscript{F} provided a baseline for our implementation of domain-aware capabilities enhanced through feature engineering to identify significant nodes and differentiate the contribution of each interaction to the affinity. In GNN\textsubscript{F}, the embedding vectors are learned simultaneously for  the protein and ligand complex as an early embedding strategy. 

These implementations enable us to leverage the vast amount of 3D structural data of both the target protein and ligands, and interactions between them which is crucial for activity, potency, and affinity prediction to accelerate in silico hit identification during early stages of drug design. The goal of our extensive study is to generalize the graph-based models by incorporating domain-aware information, features, and biophysical properties and by utilizing a large amount of data including a variety of targets. The test accuracy for GNN\textsubscript{F} reached 0.951 on a distinct sample set. We achieved top GNN\textsubscript{F} performance with the target overlap sample set, resulting in a test accuracy of 0.979 (0.958 for GNN\textsubscript{P}), providing a basis that further generalizing our model can produce top classification performance. In addition, we used the GNN\textsubscript{P} model to evaluate the performance on SARS-CoV-2 (M\textsuperscript{Pro}) and NSP15 as target proteins. The predicted probability distribution of target actives shows that GNN\textsubscript{P} can identify a high percentage of molecules with potential binding affinity against the receptor. 

Our GNN models were further modified for regression tasks to predict binding affinities and pIC\textsubscript{50} in comparison with experimentally measured values. Experimentation with regression problems such as pIC\textsubscript{50}, experimental affinity and docking binding affinity shows that our graph-based featurization of protein and ligands not only captures the binding probability but is efficient enough to learn other important factors associated with PLI. In terms of prediction, our GNN model outperform existing models \cite{fast, stepniewska2018development, jimenez2018k}  with highest prediction correlation coefficients. Using PDBbind2016 data, we achieved Pearson correlation coefficients of 0.66 and 0.65 on experimental affinity prediction and 0.50 and 0.51 on pIC\textsubscript{50} prediction using GNN\textsubscript{F} and GNN\textsubscript{P}, respectively. Even with limited 3D structural data for the pIC\textsubscript{50} dataset, we achieved comparable performance to existing methods that were trained on relatively larger 2D sequence datasets. With the availability of pIC\textsubscript{50} data and corresponding protein structure predicted from Alphafold \cite{Alphafold}, the GNN model performance can be further improved. 

Our model is unique and novel in that it considers the 3D structures of the protein and ligand to predict affinity and pIC\textsubscript{50}, which is key to  provide a quantitative measure of the potency and selectivity of a candidate to inhibit a specific protein target. To accelerate in silico hit identification and lead optimization in the early stage of drug design, our GNN\textsubscript{P} model can be used to screen a large ligand library to predict either biophysical properties or activities against a given protein target or set of targets for specific disease.
 
\section*{Acknowledgements }
This research was supported by Laboratory Directed Research and Development Program and Mathematics for Artificial Reasoning for Scientific Discovery investment at the Pacific Northwest National Laboratory, which is operated by Battelle for the U.S. Department of Energy under Contract DE-AC06-76RLO. We thank  Rajendra Joshi, Darin Hauner, and Andrew McNaughton at PNNL for discussion on the protein-ligand interactions and docking simulations. We extend our thank to Sutanay Choudhury, Khushbu Agarwal at PNNL and Aman Ahuja at Virginia Tech for extensive discussion on graph based models for accurate representation of 3D structural coordinates of protein-ligand complexes.  

\section*{Competing interests}
The authors declare no competing interests. 

\section{Supporting Information}

The Supporting information is available with Table (S1-S6) and
Figures (S1-S8)) detailing data, models, hyperparameter optimization, Performance Metrics for Classification and Regression GNN
Models, IBS molecule properties, and Docking Binding affinity Data and Results used in this study. 
Description about hyper-parameters and the trained GNN\textsubscript{P} and GNN\textsubscript{f} models and the code to reproduce this study is available at https://github.com/nkkchem/pf-gnn\_pli and  https://github.com/PNNL-CompBio/pf-gnn\_pli.

\section {Data and Code availability }
We collected and curated protein-ligand complexes from two public datasets, DUD-E \cite{mysinger2012directory} and PDBbind \cite{wang2004PDBbind, wang2005PDBbind} which are described in further detail in Dataset Preparation subsection. SARS-CoV-2 and other dataset is generated in using high throughput docking simulations and the IBS dataset is created using known synthetic compounds from the IB screening database (www.ibscreen.com). 
The trained GNN\textsubscript{P} and GNN\textsubscript{f} models  and the code to reproduce this study is available at https://github.com/nkkchem/pf-gnn\_pli and  https://github.com/PNNL-CompBio/pf-gnn\_pli.


\section*{Author Contributions}
N.K. and J.A.B. contributed concept and implementation. C.K. M.B, J.A.B and N.K. co-designed experiments and  were responsible for programming. All authors contributed to the interpretation of results and wrote the manuscript. All authors reviewed and approved the final manuscript.
\bibliography{sample}

\section{TOC Graphic}

\begin{figure}[H]
\begin{center}
    \centering
    \includegraphics[width=10cm, height=10cm, keepaspectratio]{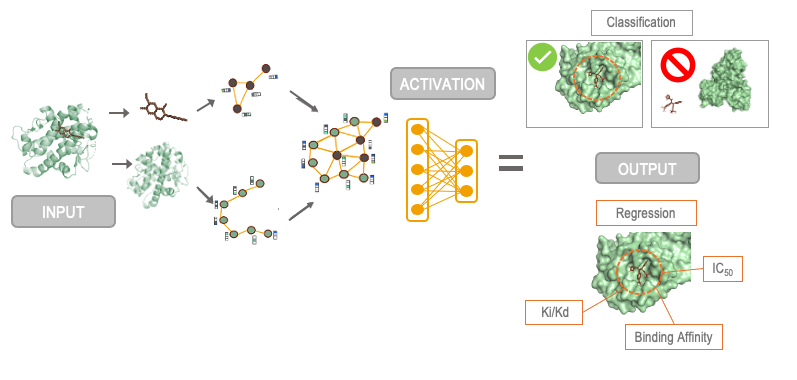}
\label{fig:TOC}
\end{center}
\end{figure}

\end{document}


\clearpage

\section{Protein Similarity Analysis}

\begin{figure}[H]
\vskip 0.2in
\begin{center}
    \centerline{\includegraphics[width=1.0\columnwidth]{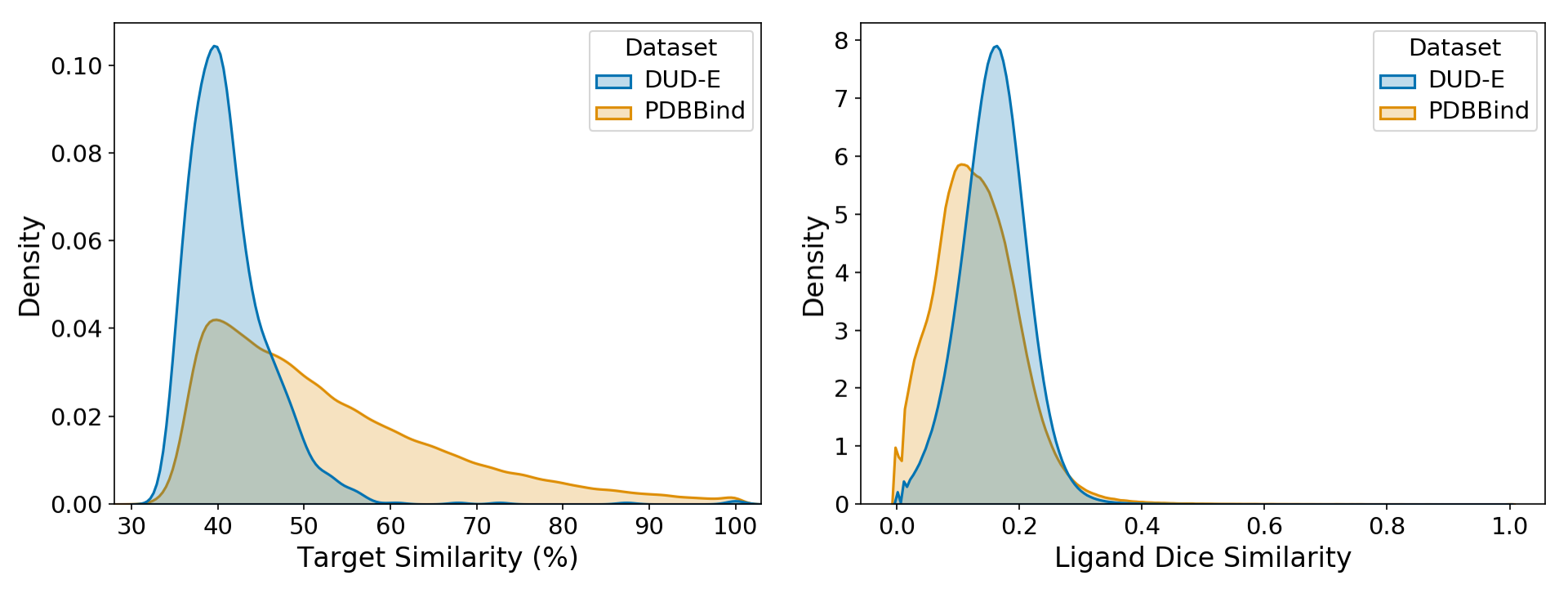}}
    \caption{Similarity of protein targets in the DUD-E and PDBbind datasets based on the homology of each target pair, and Dice similarity of the Morgan fingerprints of ligands in the DUD-E and PDBbind datasets. Because the DUD-E dataset contains a large number of ligands, not all pairs could be analyzed, so 10,000 ligands were chosen at random from the dataset for analysis.}
    \label{fig:similarity_SI}
\end{center}
\vskip -0.2in
\end{figure}

\section{Hyperparameter Optimization}

Trainings were carried out over 200 epochs on a quarter of the data taken from our dataset consisting of 2,000 samples/target with 79 targets, using the same train-test split for all trainings. We used the parameters from our best model to create a baseline: a learning rate of 0.0001, four attention heads, and a layer dimension of 70 produced an average test AUROC of 0.854. The decreased performance compared to the models discussed in the Results \& Discussion section is expected due to the reduced set of data used to train the model.
Thirty of the 36 combinations trained without error.

\begin{table}
  \caption{Values of examined hyperparameters: learning rate (lr), number ($N$) of attention heads, and dimension ($D$) of the GAT layer in each attention head. A grid search was performed on each combination of parameters. The optimal combination is shown in bold.}
  \label{tab:hypTable_SI}
  \begin{tabular}{lcc}
    \hline
lr & $N$ & $D$  \\
    \hline
0.001 & \textbf{2} & \textbf{70} \\
\textbf{0.0001} & 3 & 140 \\
0.00001 & 4 & 210 \\
 & & 280 \\
    \hline
  \end{tabular}
\end{table}

\begin{table}[H]
\vskip 0.15in 
\begin{center}
\begin{sc}
\begin{tabular}{lcc}
\toprule
HYPERPARAMETER SET & TRAIN ROC AVG & TEST ROC AVG \\
\midrule
Lr\textunderscore 0.001\textunderscore n5\textunderscore d70 & 0.754 & 0.771 \\
Lr\textunderscore 0.001\textunderscore n4\textunderscore d210 & 0.606 & 0.647 \\
Lr\textunderscore 0.001\textunderscore n4\textunderscore d140 & 0.605 & 0.661 \\
Lr\textunderscore 0.001\textunderscore n3\textunderscore d70 & 0.834 & 0.835 \\
Lr\textunderscore 0.001\textunderscore n3\textunderscore d280 & 0.610 & 0.630 \\
Lr\textunderscore 0.001\textunderscore n3\textunderscore d210 & 0.672 & 0.694 \\
Lr\textunderscore 0.001\textunderscore n3\textunderscore d140 & 0.714 & 0.742 \\
Lr\textunderscore 0.001\textunderscore n2\textunderscore d70 & 0.937 & 0.861 \\
Lr\textunderscore 0.001\textunderscore n2\textunderscore d140 & 0.788 & 0.781 \\
Lr\textunderscore 0.001\textunderscore n2\textunderscore d280 & 0.762 & 0.755 \\
Lr\textunderscore 0.001\textunderscore n2\textunderscore d210 & 0.767 & 0.773 \\
Lr\textunderscore 0.0001\textunderscore n4\textunderscore d128 & 0.898 & 0.854 \\
Lr\textunderscore 0.0001\textunderscore n3\textunderscore d70 & 0.894 & 0.855 \\
Lr\textunderscore 0.0001\textunderscore n3\textunderscore d210 & 0.930 & 0.860 \\
Lr\textunderscore 0.0001\textunderscore n3\textunderscore d140 & 0.917 & 0.862 \\
Lr\textunderscore 0.0001\textunderscore n2\textunderscore d70 & 0.912 & 0.864 \\
Lr\textunderscore 0.0001\textunderscore n2\textunderscore d210 & 0.950 & 0.861 \\
Lr\textunderscore 0.0001\textunderscore n2\textunderscore d140 & 0.940 & 0.853 \\
Lr\textunderscore 0.00001\textunderscore n4\textunderscore d70 & 0.707 & 0.726 \\
Lr\textunderscore 0.00001\textunderscore n4\textunderscore d210 & 0.776 & 0.783 \\
Lr\textunderscore 0.00001\textunderscore n4\textunderscore d140 & 0.748 & 0.769 \\
Lr\textunderscore 0.00001\textunderscore n3\textunderscore d70 & 0.735 & 0.758 \\
Lr\textunderscore 0.00001\textunderscore n3\textunderscore d210 & 0.778 & 0.772 \\
Lr\textunderscore 0.00001\textunderscore n3\textunderscore d140 & 0.763 & 0.755 \\
Lr\textunderscore 0.00001\textunderscore n2\textunderscore d70 & 0.713 & 0.749 \\
Lr\textunderscore 0.00001\textunderscore n2\textunderscore d210 & 0.776 & 0.783 \\
Lr\textunderscore 0.00001\textunderscore n2\textunderscore d140 & 0.774 & 0.782 \\
\hline
\end{tabular}
\caption{\label{tab:hypTableSI}}
\end{sc}
\end{center}
\end{table}
\clearpage
\section{Performance Metrics for Classification and Regression GNN Models}
\begin{table}[H]
    \centering
    \begin{tabular}{cccc}
    \hline
    MODEL & TARGET & SAMPLES & ACC \\
    \hline
    GNN\textsubscript{F} & 17 & 1k & 0.764 \\
    GNN\textsubscript{F} & 79 & 1k & 0.888 \\
    GNN\textsubscript{F} & 96 & 1k & 0.851\\
    GNN\textsubscript{F} & 17 & 2k & 0.764 \\
    GNN\textsubscript{F} & 79 & 2k & 0.912 \\
    GNN\textsubscript{F} & 96 & 2k & 0.900 \\
    GNN\textsubscript{P} & 17 & 1k & 0.686 \\
    GNN\textsubscript{P} & 79 & 1k & 0.837 \\
    GNN\textsubscript{P} & 96 & 1k & 0.791\\
    GNN\textsubscript{P} & 17 & 2k & 0.681 \\
    GNN\textsubscript{P} & 79 & 2k & 0.880 \\
    GNN\textsubscript{P} & 96 & 2k & 0.828 \\
    \hline
    \end{tabular}
    \caption{Accuracy of each experiment evalutating the affects of additional DUD-E targets as well as sample consideration per target for both the GNN\textsubscript{F} and GNN\textsubscript{P}}
    \label{tab:ExprAccSI}
\end{table}

\begin{table}
  \centering
\begin{adjustbox}{width=1\textwidth}
\small
  \begin{tabular}{ccccc|cccc}
    \hline

    &\multicolumn{4}{c|}{\textbf{Just core set}} &\multicolumn{4}{c}{\textbf{Entire test set}}\\
    \hline
     \textbf{Model} &\textbf{RMSE} & \textbf{MAE} & \textbf{Pearson r} & \textbf{Spearman r} & \textbf{RMSE} & \textbf{MAE} & \textbf{Pearson r} & \textbf{Spearman r}\\
\hline
 GNN\textsubscript{F}-EBA$^*$ & 1.73 & 1.42 & 0.65 & 0.63 & 1.61 & 1.28 & 0.58  & 0.57 \\
    \textbf{GNN\textsubscript{P}-EBA$^*$} & \textbf{1.73} & \textbf{1.39} & \textbf{0.62} & \textbf{0.62} & \textbf{1.55} & \textbf{1.21} & \textbf{0.58} & \textbf{0.59}  \\
    Pafnucy \cite{stepniewska2018development} & 1.42 & 1.13 & 0.78 & - & - & - & - & -\\
    SG-CNN (R + G) \cite{fast}& 1.37 & 1.08 & 0.78 & 0.76 & - & - & - & - \\
    3D-CNN (R + G) \cite{fast} & 1.68 & 1.33 & 0.67 & 0.65 & - & - & - & - \\
    midlevel fusion \cite{fast} & 1.308 & 1.019 & 0.810 & 0.807 & - & - & - & -\\   
    \textbf{K\textsubscript{DEEP}} \cite{jimenez2018k} & \textbf{1.27} & \textbf{-} & \textbf{0.82} & \textbf{0.82} & \textbf{-} & \textbf{-} & \textbf{-} & \textbf{-}\\
\hline
  \end{tabular}
  \end{adjustbox}
  \caption{Performance comparison of our GNN models in predicting Experimental affinity on PDBbind2016 core set. (*Cases where the test set contains more than just the core set)}
  \label{tab:PDBbind_2016_core_set}
\end{table}

\subsection{PDBbind2016 dataset split}
\begin{table}[H]
\begin{center}
\begin{tabular}{lccc}
\hline

Dataset & Total   & Train   & Test \\
 &  samples &  samples  &  samples \\
\hline
PDBbind2016-EBA:75:25 & 11,674 & 8,248 & 3,136 \\
PDBbind2016-EBA:80:20 & 11,674 & 8,808 & 2,276 \\
PDBbind2016-EBA:90:10 & 11,674 & 10,246 & 1,428 \\
PDBbind2016-EBA:core & 11,674 & 11,384 & 290 \\

\hline
\end{tabular}

\end{center}

\caption{Table summarizing the PDBbind2016 dataset split used for Experimental Affinity regression models}
\label{tab:table_PDBbind2016_splits}
\end{table}

The PDBbind2016-EBA core dataset consists of the entire general and refined as the train set and just the core set as the test set. The rest of the model's test set includes a fraction of the general and refined dataset along with the core set.

\centerline{\textbf{GNN\textsubscript{F} and GNN\textsubscript{P} 1k and 2k Comparison}}

\begin{figure}[H]
    \centering
    \includegraphics[width=0.60\textwidth]{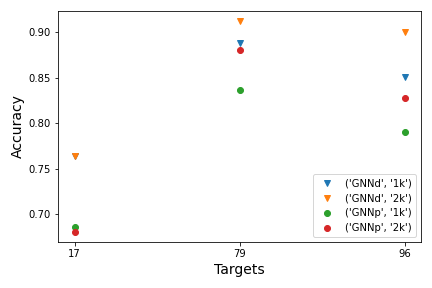}
    \caption{Dataset performance comparisons of the GNN\textsubscript{F} and GNN\textsubscript{P} models. Datasets composed of 17,79, and 96 DUD-E targets, all available PDBbind data, and implementations of 1,000 and 2,000 samples per DUD-E target. This shows how increasing the amount of targets included in the sample set, and the amount of individual samples affect model performance. }
    \label{fig:comparison}
\end{figure}

\subsection{Top-N Ranks Analysis of Classification Models}
We compare the performance of GNN with docking in identifying protein-ligand complexes in top-n ranks. We choose the best GNN model (2k\_79) to get predictions on the train and the test dataset. For the GNN, we rank the models in ascending order of predicted binding probabilities. For docking we rank the models based on the descending order of binding affinity scores obtained as a part of docking pipeline. We assess the performance of each of the scoring method (GNN and docking) on their ability to identify protein-ligand models with RMSD less than 2Å from the experimental crystal structure. Since we had crystal structure available only for PDBbind data, we performed the analysis on just the PDBbind dataset, once on the PDBbind test targets and once on the entire PDBbind data. Here we report the percentage of targets identified which had ant least one model identified in top-N ranks when ranked according to the GNN and Docking scores. Here, we are evaluating on 676 targets from training and 159 targets from test  which were commong for all the three scoring methods.From the graph. We can see that GNN\textsubscript{F} and docking gave a nearly same performance on training data while GNN\textsubscript{P}  has been the poor performer. The performance on the test data is same for all the methods.

\begin{figure}[H]

\begin{center}
    \centerline{
    \includegraphics[width=0.5\textwidth]{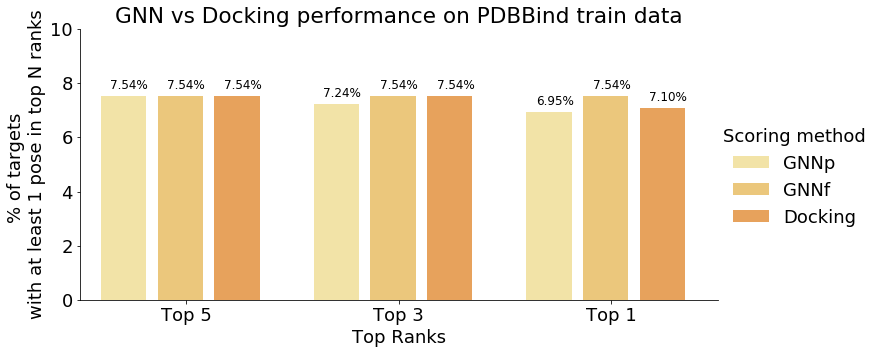}%
    \includegraphics[width=0.5\textwidth]{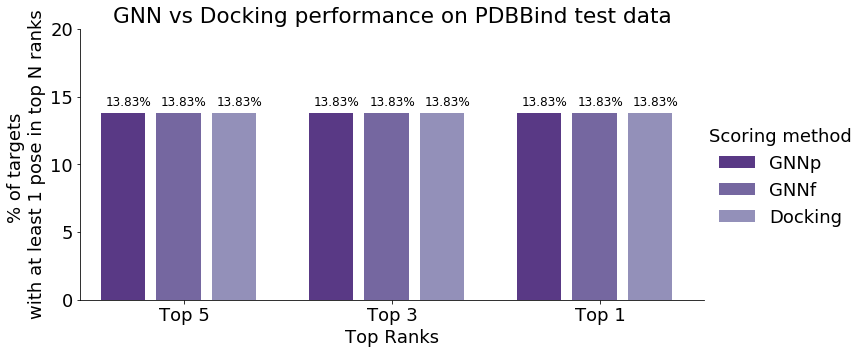}}
    \caption{Comparision of GNN\textsubscript{p} and GNN\textsubscript{f} model with Docking. Each bar corresponds to the percentage of protein-ligand complexes with at least 1 docked pose in Top-N ranks which have an RMSD less than 2\AA~from it's crystal structure. }
    \label{fig:top_n_ranks_docking}
\end{center}
\end{figure}

To measure the potency of GNN models. We assess the model's ability to identify docked poses per protein-ligand complex. With this, we not only measure the ability of the model to identify an active protein-ligand pose but also its ability to identify the best pose amongst multiple docked poses for a protein-ligand complex (see Fig. S\ref{fig:top_n_ranks_docking}).

\subsection{IBS molecule properties}

\begin{figure}[H]

        \includegraphics[width=0.33\textwidth]{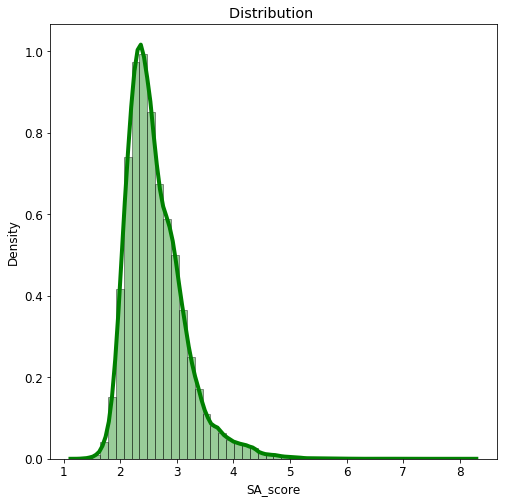}%
        \includegraphics[width=0.33\textwidth]{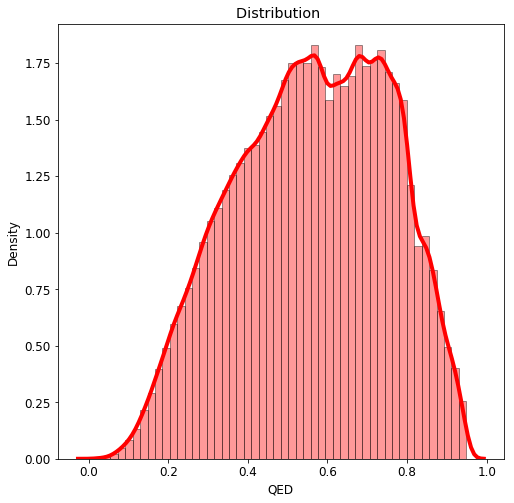}
        \includegraphics[width=0.33\textwidth]{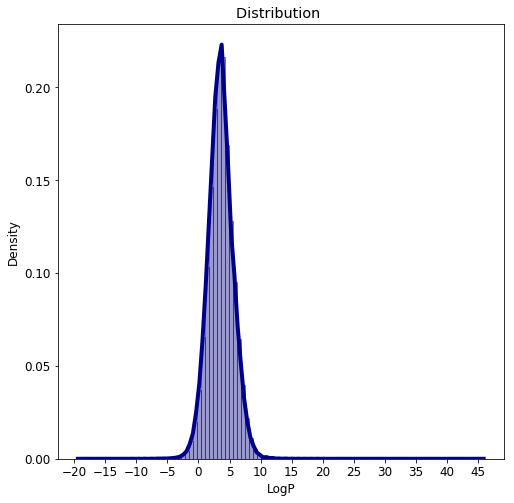}
            \caption{Distribution of properties of IBS molecules to measure their drug-likeliness. The properties include the synthetic accessibility (SA) score, quantitative estimation of drug-likeliness (QED), and the partition coefficient (logP).}
            \label{fig:IBSprop}

\end{figure}

The SA score is the synthesizability of generated molecules and ranges between 0-10, where the lower end suggests increased accessibility. QED is a measures the quantifying and ranking the drug-likeness of a compound. The values range from 0 for unfavorable to 1 for favorable. The partition coefficient, logP is a measure that determines the physical nature of a compound and its ability to reach the target in the body. A positive logP value indicates the compound is lipophilic and a negative logP value indicates a hydrophilic compound. We can see that more than 60\% of compounds have a low SA score ( indicates easily synthesizable) and a high QED ( indicates high drug-likeliness) and a lopP between -0.4 and 5.6. These properties suggest that considerably a greater number of IBS molecules could have a high potency to bind to a known receptor.

\subsection{IBS molecules - Binding probability vs molecular weight}

\begin{figure}[H]

\begin{center}
\centerline{
\includegraphics[width=0.50\textwidth]{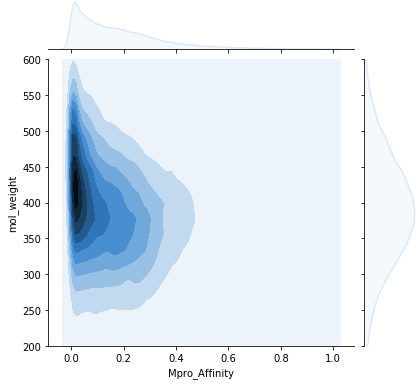}%
\includegraphics[width=0.50\textwidth]{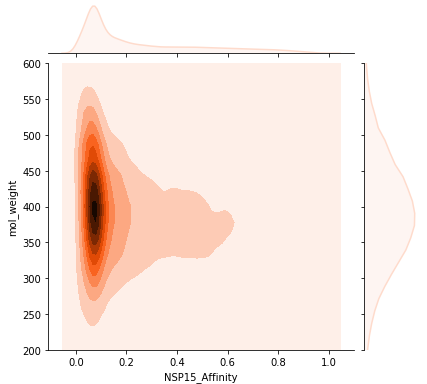}}
    \caption{Density plot for molecular weight vs binding probability for IBS molecules}
    \label{fig:IBS_mol_weight}
\end{center}

\end{figure}

\subsection{pIC\textsubscript{50} of Active Targets and Predicted Binding Affinity Relationship Analysis}
\label{subsection:ic50}
In this section we would discuss about the IC\textsubscript{50} activity of the DUD-E active targets and their correlation with the predicted binding probability. We chose the predicted binding probability of the model with lowest docking energy since we had multiple poses for each molecule obtained as a part of the docking pipeline.
We found that majority of the molecules have pIC\textsubscript{50} between 10 and 20. The expected trend here is to have high binding probability associated with a high pIC\textsubscript{50} value.

\begin{figure}[!ht]
  \centering
  \subfloat{\includegraphics[width=.4\textwidth]{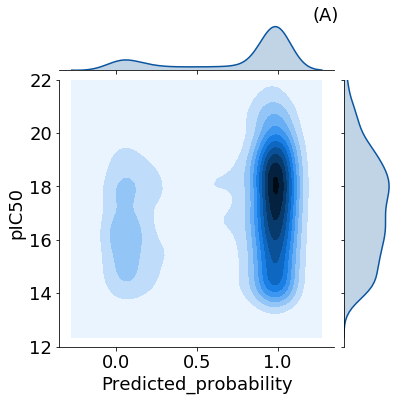}}\quad
  \subfloat{\includegraphics[width=.4\textwidth]{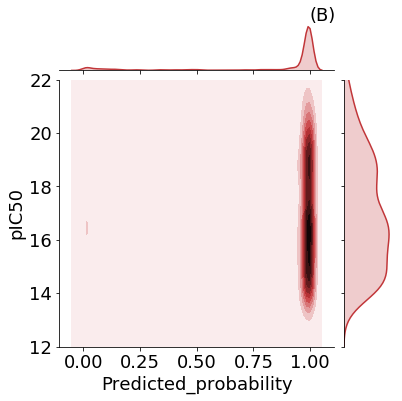}}\\
  \subfloat{\includegraphics[width=.45\textwidth]{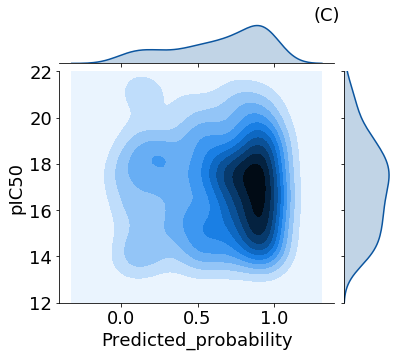}}\quad
  \subfloat{\includegraphics[width=.4\textwidth]{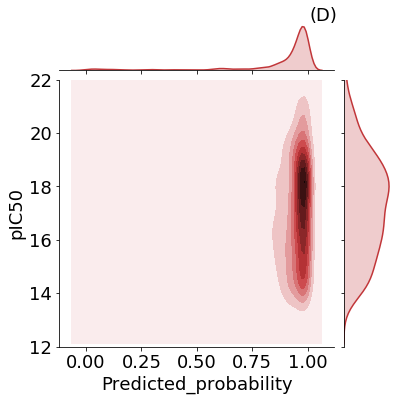}}
  \caption{pIC\textsubscript{50} vs Binding probability for DUD-E active molecules. Figure (A) and (B) correspond to test and train for GNN\textsubscript{F} and figure (C) and (D) correspond to the GNN\textsubscript{P} model. The model was able to associate high binding probabilities to high  pIC\textsubscript{50} valued molecules.}
  \label{fig:pic50_binding_probability}
\end{figure}

The GNN\textsubscript{F} model displays increased performance to the GNN\textsubscript{P} as  
  the density of higher pIC\textsubscript{50} valued molecules is concentrated at high binding probability values, as shown in Figure \ref{fig:pic50_binding_probability}. We can see a similar trend for GNN\textsubscript{P} model as well, except that the model identifies quite a majority of high pIC\textsubscript{50} valued compounds as poorly binding complexes. While looking at the performance on the training dataset, both GNN\textsubscript{F} and GNN\textsubscript{P}
 showed a good learning by attributing high pIC\textsubscript{50} compounds with a high binding probability. Overall we can say that GNN\textsubscript{F}  has shown a good performance on both training and test set while learning the corellation between the experimental binding affinity and probability of binding. 

GNN\textsubscript{p} test set has a bimodal distribution for probability 0, so it gave errors on a number of low IC50 samples compared to GNN\textsubscript{f} where the errors were mainly for high IC50 samples 

\subsection{GNN vs FAST Model Comparison}
\begin{table}[H]
    \centering
    \begin{tabular}{cccccc}
    \hline
    MODEL & RMSE & MAE & Pearson r & Spearman r & $r^2$\\
    \hline
    GNN\textsubscript{F}-EBA (R) & 1.92 & 1.56 & 0.51 & 0.50 & 0.26  \\
    GNN\textsubscript{F}-EBA (G) & 1.84 & 1.48 & 0.58 & 0.59 & 0.34  \\
    GNN\textsubscript{F}-EBA-(R + G) & 1.75 & 1.42 & 0.61 & 0.60 & 0.37  \\
    GNN\textsubscript{P}-EBA (R) & 1.72 & 1.37 & 0.63 & 0.62 & 0.40  \\
    \textbf{GNN\textsubscript{P}-EBA (G)} & \textbf{1.66} & \textbf{1.33} & \textbf{0.66} & \textbf{0.67} & \textbf{0.43}  \\
    GNN\textsubscript{P}-EBA-(R + G) & 1.69 & 1.33 & 0.64 & 0.66 & 0.41  \\
    SG-CNN (R) & 1.65 & 1.32 & 0.66 & 0.64 & 0.42  \\
    SG-CNN (G) & 1.50 & 1.19 & 0.74 & 0.74 & 0.51  \\
    \textbf{SG-CNN (R + G)} & \textbf{1.37} & \textbf{1.08} & \textbf{0.78} & \textbf{0.76} & \textbf{0.60}  \\
    3D-CNN (R) & 1.5 & 1.16 & 0.72 & 0.71 & 0.52  \\
    3D-CNN (G) & 1.65 & 1.29 & 0.64 & 0.65 & 0.42  \\
    3D-CNN (R + G) & 1.68 & 1.33 & 0.67 & 0.65 & 0.39  \\
  \hline
    \end{tabular}
    \caption{Performance comparison of our GNN models and FAST \cite{fast} models on the same data split from PDBbind2016. While the SG-CNN shows the best performance on the R + G split and 3D-CNN gives best performance on the refined set, we achieve the best performance using the GNN\textsubscript{P} model on the general set}
    \label{tab:FAST_GNN_same_split}
\end{table}

\subsection{Docking Binding affinity Data and Results}

For Docking Binding Affinity (DBA) regression models, we used 1000 positive and 1000 negative poses per target in DUD-E dataset as we did for the classification dataset. In addition, for the DBA scores, we used the docking score as obtained from the docking pipeline for all the PDBbind and DUD-E decoys.
\begin{table}[H]
\begin{center}
\begin{small}

\resizebox{\textwidth}{!}{\begin{tabular}{lccccccc}
\hline
Dataset & Total  & Total   & Train   & Test  & Train  & Test  & Total  \\
 &  targets &  ligands  &  targets  &  targets & samples &  samples &  samples \\
\hline

PDBbind2018-DBA  & 991 & 991 & 867 & 124 & 5,410 & 812 & 6,222 \\
DUD-E-DBA &  65 & 5,358 & 54 & 11 & 105,814 & 22,218 & 128,032 \\

\hline
\end{tabular}}

\end{small}

\end{center}

\caption{Table summarizing the dataset used for Docking Binding Affinity (DBA) regression models} 
\label{tab:DBA_dataset}
\end{table}

\begin{table}[H]
    \centering
    \begin{tabular}{cccccc}
    \hline
    MODEL & RMSE & MAE & Pearson r & Spearman r & $r^2$\\
    \hline
    GNN\textsubscript{F}-DBA &  0.80 & 0.63 & 0.78 & 0.79 & 0.62 \\
    GNN\textsubscript{P}-DBA & 0.84 & 0.66 & 0.72 & 0.70 & 0.51 \\
    \hline
    \end{tabular}
    \caption{Performance comparison of our GNN models in predicting Docking affinity on test dataset. }
    \label{tab:DBA_results}
\end{table}